\documentclass{article}
\usepackage{ijcai23}

\usepackage{times}
\usepackage{soul}
\usepackage{url}
\usepackage[hidelinks]{hyperref}
\usepackage[utf8]{inputenc}
\usepackage[small]{caption}
\usepackage{graphicx}
\usepackage{amsmath}
\usepackage{amsthm}
\usepackage{booktabs}
\usepackage{algorithm}
\usepackage{algorithmic}
\usepackage[switch]{lineno}
\usepackage{multirow}
\usepackage{multicol}
\usepackage{float}
\usepackage{subfig}
\usepackage{bm}
\usepackage{pifont}
\usepackage{amssymb}

\title{Attribute Localization and Revision Network for Zero-Shot Learning}

\author{
Junzhe Xu$^1$
\and
Suling Duan$^2$\and
Chenwei Tang$^1$\and
Zhenan He$^1$\thanks{Corresponding Author.}\And
Jiancheng Lv$^1$
\affiliations
$^1$College of Computer Science, Sichuan University, China\\
$^2$School of Statistics, Chengdu University of Information Technology, China\\
\emails
druryxu@stu.scu.edu.cn,
dsl@cuit.edu.cn,
\{tangchenwei, zhenan, lvjiancheng\}@scu.edu.cn
}

\begin{document}

\maketitle

\begin{abstract}
   Zero-shot learning enables the model to recognize unseen categories with the aid of auxiliary semantic information such as attributes. Current works proposed to detect attributes from local image regions and align extracted features with class-level semantics. In this paper, we find that the choice between local and global features is not a zero-sum game, global features can also contribute to the understanding of attributes. In addition, aligning attribute features with class-level semantics ignores potential intra-class attribute variation. To mitigate these disadvantages, we present Attribute Localization and Revision Network in this paper. First, we design Attribute Localization Module (ALM) to capture both local and global features from image regions, a novel module called Scale Control Unit is incorporated to fuse global and local representations. Second, we propose Attribute Revision Module (ARM), which generates image-level semantics by revising the ground-truth value of each attribute, compensating for performance degradation caused by ignoring intra-class variation. Finally, the output of ALM will be aligned with revised semantics produced by ARM to achieve the training process. Comprehensive experimental results on three widely used benchmarks demonstrate the effectiveness of our model in the zero-shot prediction task.
\end{abstract}

\section{Introduction}
Human beings have an outstanding ability to recognize objects that have never been seen before. However, traditional deep learning models generally struggle to identify classes that do not appear in the training set, which largely limits deep learning's ability to simulate human behavior. To tackle this problem, Zero-Shot Learning (ZSL) \cite{Lampert2009Learning} has been proposed to identify images from unseen classes. 

The key to ensuring the success of ZSL is the semantics of all classes provided during training, which acts as a bridge to help the model transfer knowledge from seen to unseen classes. In a typical setting \cite{Xian2018ZeroShot}, a semantic is a vector with the value at each dimension representing the saliency value of a certain attribute, i.e., if an attribute does not present on the image, the value will be zero. 

There are two mainstream ZSL settings, conventional ZSL \cite{Lampert2009Learning} and generalized ZSL (GZSL) \cite{Chao2016An}. The former assumes that all samples in the test set belong to unseen classes. However, samples belonging to seen classes may also appear during testing in real-world scenarios. Hence, GZSL has been introduced. It adds samples from seen classes into the test set, which enlarges the classification space from unseen classes to both seen and unseen classes.

\begin{figure}[t]
  \centering
  \includegraphics[width=1\linewidth]{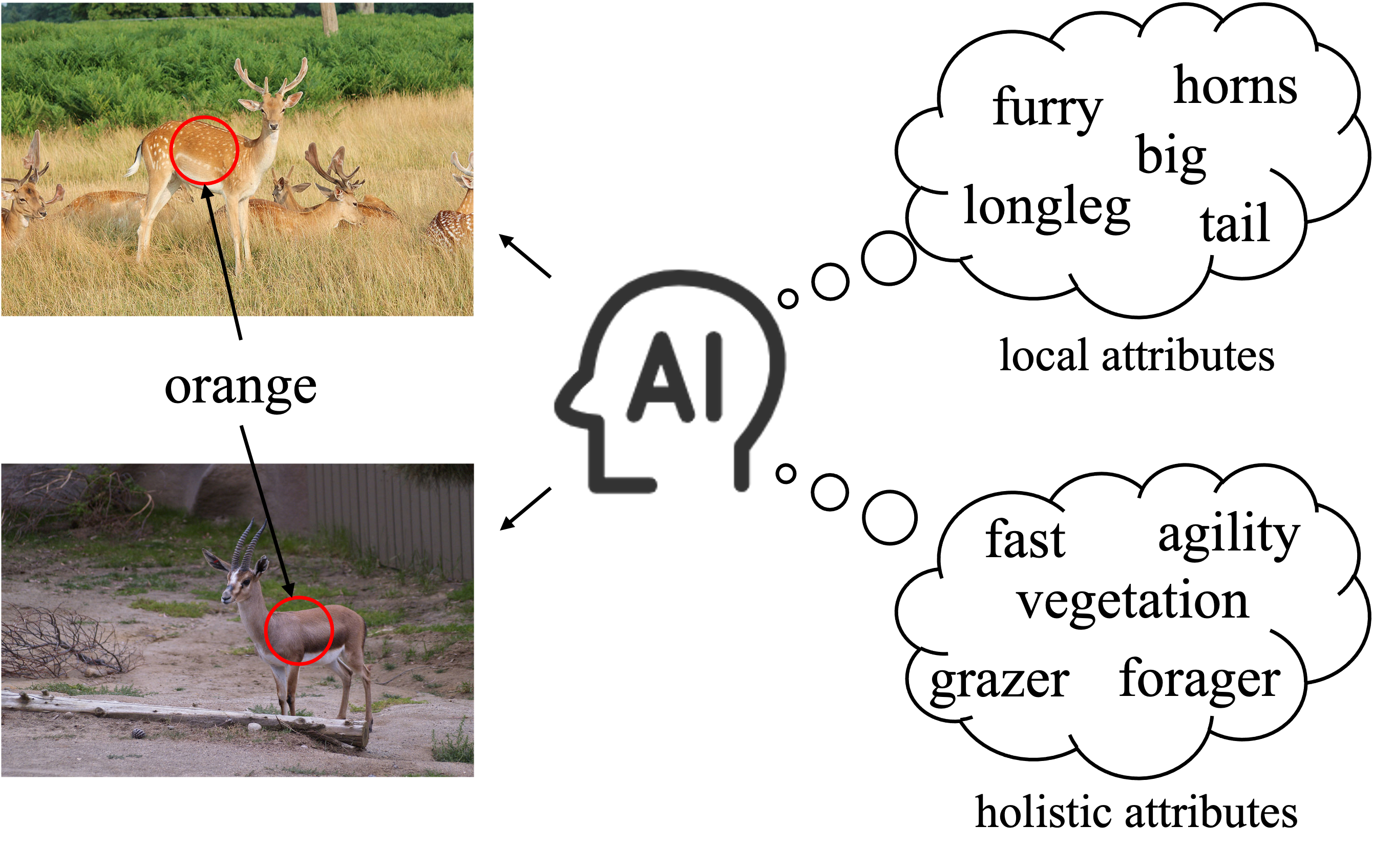}
  \caption{Two existing problems an attention-based model will encounter during training, i.e., potential attribute variation within a class (left side) and two different scales of attributes (right side).}
  \label{introfig}
\end{figure}

In the beginning, GZSL follows the idea of conventional ZSL to leverage cross-modal projection, which aligns visual and semantic features to build a visual-semantic space, and predictions are made through the nearest neighbor algorithm. Nevertheless, since the model is fully trained on seen classes, it will be severely biased towards seen classes during testing, which tends to classify unseen samples into seen classes. Since such an overfitting problem is caused by the extreme data imbalance between seen and unseen classes, generation-based methods \cite{Xian2018Feature,Chen2021FREE} proposed to directly generate unseen samples and convert the GZSL task to a fully supervised task. However, generation-based methods use the global image representations extracted by a pre-trained backbone as training samples, where discriminative local information will be submerged in the global representation. To this end, attention-based methods \cite{Xu2020Attribute,Chen2021TransZero,Chen2022MSDN} abandoned utilizing global features but proposed to locate and evaluate the saliency value of each attribute directly from image regions, which avoids the loss of local discriminative features.
 
The key to success in the attention-based method is whether it can accurately establish a one-to-one correspondence between image regions and attributes. Thus, previous works concentrate on building better local representation extractors to capture discriminative attribute information. However, there may be no clear clues in the image for building such a correspondence. To illustrate, there exist attributes that are more semantically related rather than visually related to the image. For a better understanding, the right side of Figure \ref{introfig} lists a few examples of different kinds of attributes that belong to the antelope. As we can see, the visually related attributes, i.e., local attributes, clearly depict a part of the image, and semantically-related attributes, i.e., holistic attributes, are the condensation of highly abstract knowledge that cannot be captured purely by eyes. In some situations, annotating local attributes manually is hard or expensive to achieve and off-the-shelf external information, e.g., knowledge graph, provides a convenient way to annotate attributes automatically. However, such external knowledge will unavoidably bring some semantically related attributes. Therefore, it is crucial to consider the existence of holistic attributes in many real applications. 

Another phenomenon that hinders the model localize attributes from the image is intra-class attribute variation, which means the same attribute will present differently in different images. We exhibit such a variation in the left side of Figure \ref{introfig}. Since two pictures are taken under different conditions, the attribute "orange" representation is diverse. Current attention-based methods adopted a single semantic vector as the ground truth learning target for all images in one class, which ignores the possible attribute variation. Evaluating the value of each attribute independently causes the individual attribute prediction result to have an effect on the overall result, so the attribute variation during training needs to be meticulously considered. 

The above two phenomena are not well addressed in previous works, which makes them two challenges that cannot be ignored on the road of ZSL development. To alleviate these problems, we propose a novel framework called Attribute Localization and Revision Network (ALRN). We first discover the importance of global features in understanding attributes rather than purely adopting local features. Since it is unreasonable to enforce the model to detect undetectable holistic attributes, the incorporation of global features helps the model understand holistic attributes from the perspective of the entire image. Also, it can act as a complement to the attention mechanism such as deducing the attribute "horns" if we know there is an antelope in the image. Therefore, our proposed Attribute Localization Module (ALM) extracts local features through the attention mechanism and global features via average pooling. Then, we introduce Scale Control Unit (SCU) to fuse these two kinds of features. To mitigate the problem of ignoring intra-class attribute variation, we propose the Attribute Revision Module (ARM). For each input image, it first generates a specific set of revision weights for all attributes and conducts an element-wise product between revision weights and class-level attribute values to generate new sets of image-level attribute values. To this end, ARM considers intra-class attribute variation from the perspective of learning from revised targets. 

The contributions of our work are summarized as follows: 

\begin{itemize}
    \item We exploit the importance of global features in understanding attributes, and the scale control mechanism is designed to aggregate local and global features.
    \item Considering possible attribute variation lies among images within the same class, we propose the attribute revision mechanism to revise class-level attribute saliency value into image-level. 
    \item The experimental results on three ZSL benchmarks demonstrate our proposed ALRN has a remarkable ability in both conventional ZSL and GZSL prediction tasks.
\end{itemize}

\section{Related Works}\label{relatedworks}
Since samples and semantics lie on different modal spaces, projection-based methods \cite{Chao2016An,Akata2015Label,Akata2015Evaluation,Frome2013DeViSE} proposed to project and align them on a common space. However, after GZSL becomes the research emphasis of ZSL, projection-based methods suffer from severe bias toward seen classes. To alleviate this problem, calibrated stacking mechanism \cite{Chao2016An}, knowledge transfer \cite{Jiang2019Transferable,Liu2018Generalized,Li2019Rethinking}, episode-training \cite{Yu2020Episode} are introduced.

Another way to prevent the model from bias toward seen classes is generating unseen samples using generative models \cite{Goodfellow2014Generative,Kingma2014AutoEncoding}, where researchers \cite{Narayan2020Latent,Xian2019FVAEGAND2,Xian2018Feature,Mishra2018AGenerative,Verma2018Generalized} build a generator with semantics as conditional information. To promote the quality of generated samples, some works try to learn the detailed distribution of seen samples \cite{Li2019Leveraging,Xie2021Generalized}, incorporate the distribution of semantics into the learning process \cite{Vyas2020Leveraging,Tang2021ZeroShot}, and add distribution regulation to synthesized seen samples \cite{Han2020Learning,Han2021Contrastive}. 


Besides projection and generation methods, researchers have turned their sights to directly localizing attributes on images in recent years. Along this line, attention mechanism \cite{Huynh2020FineGrained,Chen2022MSDN,Chen2021TransZero,Liu2021GoalOriented,Liu2019Attribute,Xie2019Attentive}, max pooling \cite{Yang2021On,Xu2020Attribute} are exploited to capture attributes from the feature map produced by a backbone.

Several works propose image-adaptive semantics \cite{Liu2019Attribute,Chou2020Adaptive} in order to solve the performance degradation brought by semantic ambiguation. However, these methods generate attention weights across all attributes and therefore are essentially class-level attention. In our method, we further fine-grain such an operation into attribute-level, where the weights for each attribute are produced independently. This modification better integrates with the goal of attribute localization to achieve a true attribute revision. Besides, our method first introduces the beneficial role of global features in attention-based methods, which further enhances the ability of our model in extracting attribute features.

\section{Proposed Method}\label{proposedmethod}
\begin{figure}[t]
  \centering
  \includegraphics[width=1\linewidth]{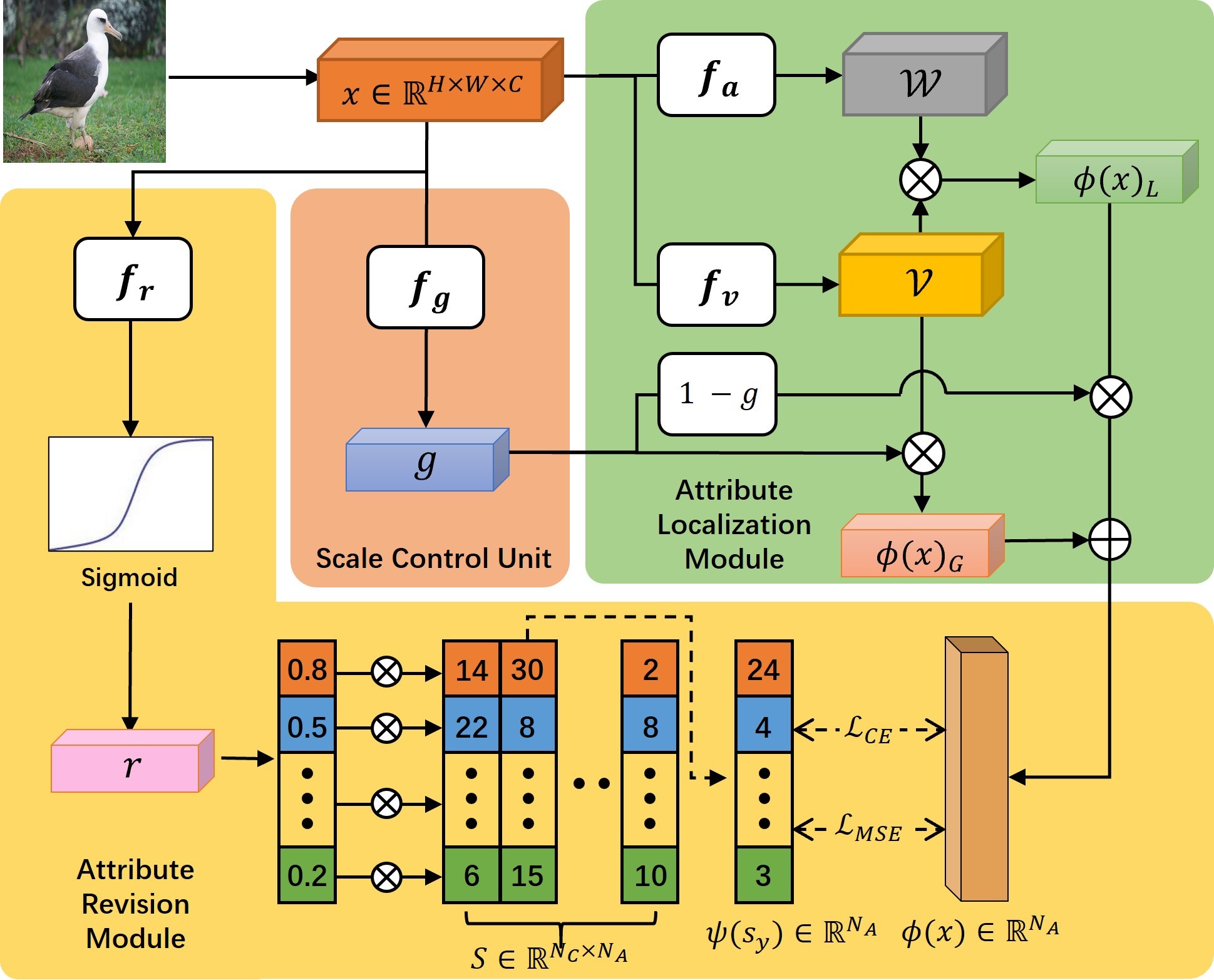}
  \caption{The framework of proposed ALRN, where $\otimes$ denotes element-wise product, and $\oplus$ is addition operation. The attribute Localization Module (green part) tends to localize and evaluate the saliency value of each attribute. The attribute Revision Module (yellow part) produces revision weight for each attribute.}
  \label{framework}
\end{figure}

In this paper, we propose a novel model called Attribute Localization and Revision Network (ALRN), which consists of two modules. The attribute Localization Module is used to find local regions of an image containing a certain attribute and incorporate the feature of the whole image as a global feature of an attribute. In addition, the proposed Scale Control Unit dynamically controls the network to concentrate on local and global features, which helps the model capture attributes in different scales. Finally, Attribute Revision Module plays a role in independently revising the ground-truth saliency value of each attribute and thus forming new image-level semantics. The whole framework can be found in Figure \ref{framework}. 

\subsection{Task Definition and Notations}
The seen dataset is denoted as $\mathcal{D}_{tr}=\{(x, y)|x\in \mathcal{X}_s, y\in \mathcal{Y}_s\}$, where $\mathcal{X}_s$ represents a collection of seen visual images, $\mathcal{Y}_s$ are the labels of $\mathcal{X}_s$. Analogously, the unseen dataset is denoted as $\mathcal{D}_{ts}=\{(x, y)|x\in \mathcal{X}_u, y\in \mathcal{Y}_u\}$, where $\mathcal{X}_u$ represents a collection of unseen visual images, $\mathcal{Y}_u$ are the labels of $\mathcal{X}_s$. It is worth noting that the seen label space and the unseen label space are disjoint, i.e., $\mathcal{Y}_s\cap \mathcal{Y}_u=\varnothing$. During the training process, semantics $\mathcal{S}\in \mathbb{R}^{N_A\times N_C}$ of all classes are provided, where $N_C$ is the total number of classes and $N_A$ represents the total number of attributes. The semantic for class $i$ can be written as $s_i=[a_i^{1},a_i^{2},...,a_i^{N_A}]^{\top}$, where $a_i^{j}$ means the saliency value of $j$th attribute with respect to class $i$. For each image, the feature map obtained through the backbone is represented as $x \in \mathbb{R}^{C\times H\times W}$.

\subsection{Attribute Localization and Scale Control}
The Attribute Localization Module (ALM) aims at positioning each attribute on an image. It first produces a set of attention weights to localize each attribute. To be specific, a $1\times 1$ convolutional layer $f_a$ is applied to convert the channel dimension of the feature map from $C$ to $N_A$, then softmax is adopted to encode the attention weight of each attribute:
\begin{align}
  \mathcal{W}       &= \text{softmax}(f_a(x)),
\end{align}
where $\mathcal{W}\in \mathbb{R}^{N_A\times H\times W}$ denotes the attention weights of all attributes. Besides attribute localization, evaluating the saliency value of each attribute is also important for ALM. Another $1\times 1$ convolutional layer $f_v$ is introduced to obtain the saliency map $\mathcal{V}$. We perform a element-wise product between attention weight map $\mathcal{W}$ and saliency map $\mathcal{V}$ to get the predicted semantic $\phi(x)_L\in \mathbb{R}^{N_A}$:
\begin{align}
  \mathcal{V}     &= f_v(x), \\
  \phi_L(x)^n &= \sum^H_{i=1}\sum^W_{j=1}\mathcal{W}^n_{i,j}\times \mathcal{V}^n_{i,j},
\end{align}
where $\phi_L(x)^n$ is the predicted saliency value for the $n$th attribute, $\mathcal{V}^n_{i,j}$ represents the value of the $n$th attribute with respect to position $(i,j)$. Note that $\mathcal{W}$ endows different attention weights to different spatial positions, we add the subscript $L$ to $\phi(x)$ to indicate that it represents local feature.

Since holistic attributes are hard to be visually detected on the image. We directly exploit the feature of the whole image to help the model understand them. The global average pooling is incorporated to generate global feature $\phi_G(x)\in \mathbb{R}^{N_A}$: 
\begin{equation}
  \phi_G(x)^n=\frac{1}{H\times W}\sum^H_{i=1}\sum^W_{j=1}\mathcal{V}^n_{i,j},
\end{equation}
where $\phi_G(x)^n$ denotes the saliency value for the $n$th attribute. 

Once local feature $\phi_L(x)$ and global feature $\phi_G(x)$ are obtained, we aggregate them together to acquire the final predicted description $\phi(x)$. To achieve this goal, we propose the Scale Control Unit (SCU) which produces a gate vector $g\in \mathbb{R}^{N_A}$ to help the model adaptively fuse the local or global feature:
\begin{align}
  \label{gate} g^n         &= \text{sigmoid}(\frac{1}{H\times W}\sum^H_{i=1}\sum^W_{j=1}f_g(x)^n_{i,j}),
\end{align}
where $f_g$ denotes $1\times 1$ convolutional layer, superscript $n$ means the value at $n$th channel dimension, and subscript $i,j$ stands for the value at spatial position $(i,j)$. The final predicted description $\phi(x)\in \mathbb{R}^{N_A}$ is generated as follows:
\begin{equation}
  \phi(x) = g\times \phi_L(x) + (1 - g)\times \phi_G(x).
\end{equation}
The predicted semantic description $\phi(x)$ incorporates both local and global information, making it able to evaluate the saliency value of both local and holistic attributes. Traditionally, $\phi(x)$ is used to align with the corresponding class-level semantic description. However, as we illustrated before, it ignores the differences between images within the same class. Therefore, we propose Attribute Revision Module to alleviate this problem.

\subsection{Attribute Revision}
Similar to ALM, the Attribute Revision Module (ARM) also need to obtain the feature of each attribute independently to let the model know how to revise the saliency value of each attribute. The revision vector $r\in \mathbb{R}^{N_A}$ is calculated as follows:
\begin{align}
  r^n      &= \text{sigmoid}(\frac{1}{H\times W}\sum^H_{i=1}\sum^W_{j=1}f_r(x)^n_{i,j}),\label{sigmoid}
\end{align}
where $f_r$ denotes $1\times 1$ convolutional layer, superscript $n$ means the value at $n$th channel dimension, and subscript $i,j$ stands for the value at spatial position $(i,j)$. A low weight value means the corresponding attribute saliency value should be suppressed. Conversely, a weight close to one means that the corresponding saliency value should be retained.


Based on the illustration above, $r$ does a element-wise product with each semantic vector $s\in \mathbb{R}^{N_A}$ in $\mathcal{S}$:
\begin{equation}
    \psi_x(s_i)=[r^1\times a^1_i, r^2\times a^2_i,...,r^{N_A}\times a^{N_A}_i]^{\top},
\end{equation}
where $\psi_x(s_i)$ means the image-level semantic vector for class $i$ with respect to image $x$, $r^i$ represents the revision weight for $i$th attribute, and $a^i_j$ is the real saliency value of the $i$th attribute for class $j$.

\subsection{Training Algorithm}
Once predicted semantic description $\phi(x)$ and image-level semantic vectors $\psi_x(s)$ are obtained, two loss functions are introduced to train the model. The first one is a cross-entropy loss which pushes $\phi(x)$ close to the target semantic vector $\psi_x(s_y)$ and far away from semantic vectors belonging to other classes:
\begin{equation}\label{celoss}
    \mathcal{L}_{CE}=-\frac{1}{N_B}\sum^{N_B}_{i=1}\log\frac{\tau\exp(\cos(\phi(x_i), \psi_{x_i}(s_y)))}{\sum_{\hat{y}\in \mathcal{Y}_s}\exp(\tau\cos(\phi(x_i), \psi_{x_i}(s_{\hat{y}})))},
\end{equation}
where $\tau$ is a scaling coefficient, $N_B$ represents the size of a mini-batch, $\psi_{x_i}(s_{\hat{y}})$ denotes the image-level semantic description of class $\hat{y}$ with respect to the $i$th image in mini-batch, and $\cos(\cdot)$ means cosine similarity. 

It is worth noting that $\mathcal{L}_{CE}$ acts as a class-level alignment between output features and semantics of different classes. However, since our model predicts the saliency value of each attribute independently, it is essential to directly align the attribute feature. Therefore, a mean squared error loss function is applied to serve as an auxiliary to $\mathcal{L}_{CE}$, which is an attribute-level alignment between the values of $\phi(x)$ and $\psi_x(s)$ in each dimension:
\begin{equation}
    \mathcal{L}_{MSE}=\frac{1}{N_B}\sum^{N_B}_{i=1}||\phi(x_i)-\psi_{x_i}(s_y)||_2^2.
\end{equation}
Therefore, the complete loss function can be written as:
\begin{equation}\label{loss}
    \mathcal{L}=\mathcal{L}_{CE} + \lambda \mathcal{L}_{MSE}.
\end{equation}
where $\lambda$ is a manually controlled hyper-parameter.

The training algorithm contains two stages, i.e, the kernel pretraining stage and the end-to-end training stage. During the kernel pretraining stage, only four sets of convolutional kernels will be trained. The number of training epoch $N_{pre}$ for this stage is a hyper-parameter tuned manually. After $N_{pre}$ training epochs, the training strategy converts to end-to-end training updating the parameter of the whole network simultaneously.

\subsection{Prediction Method}
In conventional ZSL, test images are only from unseen classes, so the classification space is limited to $\mathcal{Y}_u$. The prediction results are calculated as follows:
\begin{equation}
    \tilde{y}=\mathop{\text{argmax}}\limits_{\hat{y}\in \mathcal{Y}_u}\cos(\phi(x),\psi(s_{\hat{y}})),
\end{equation}
where $\tilde{y}$ is the predicted label for input image $x$ and $s_{\hat{y}}$ represents the semantic vector of class $\hat{y}$. For GZSL, to prevent the model from overly concentrating on seen class, the calibration stacking mechanism \cite{Chao2016An} is applied to reduce prediction scores on seen classes.
\begin{equation}
    \tilde{y}=\mathop{\text{argmax}}\limits_{\hat{y}\in \mathcal{Y}_s\cup \mathcal{Y}_u}(\tau\cos(\phi(x),\psi_x(s_{\hat{y}})) - \mu\mathbb{I}[\hat{y}\in \mathcal{Y}_s]),
\end{equation}
where $\mathbb{I}[\cdot]$ is an indication function whose value will be one if $\hat{y}\in \mathcal{Y}_s$, $\mu$ is a reduction factor needing to be tuned manually, and $\tau$ is the same hyper-parameter in Eq. \ref{celoss}.

\begin{table*}[ht]
		\centering
		\resizebox{1.0\linewidth}{!}{
			\begin{tabular}{c|c|c|ccc|ccc|ccc}
				\toprule
				&\multirow{2}{*}{\textbf{Methods}}&\multirow{2}{*}{\textbf{Venues}} 
				&\multicolumn{3}{c|}{\textbf{CUB}}&\multicolumn{3}{c|}{\textbf{SUN}}&\multicolumn{3}{c}{\textbf{AWA2}}\\
				&&& \bm{$S$} & \bm{$U$} & \bm{$H$} & \bm{$S$} & \bm{$U$} & \bm{$H$} & \bm{$S$} & \bm{$U$} & \bm{$H$} \\
				\midrule
                \multirow{8}{*}{\rotatebox{90}{\textbf{Projection}}}
				&DeViSE \cite{Frome2013DeViSE} & NIPS'13 & 53.0 & 23.8 & 32.8 & 16.9 & 20.9 & 23.1 & 74.7 & 17.1 & 27.8 \\
				&SJE \cite{Akata2015Evaluation} & CVPR'15 & 59.2 & 23.5 & 33.6 & 30.5 & 14.7 & 19.8 & 73.9 & 8.0 & 14.4 \\
				&DCN \cite{Liu2018Generalized} & NIPS'18 & 60.7 & 28.4 & 38.7 & 37.0 & 25.5 & 30.2 & -- & -- & -- \\
				&TCN \cite{Jiang2019Transferable} & ICCV'19 & 48.1 & 33.3 & 39.4 & 18.5 & 30.9 & 23.1 & 72.8 & 52.1 & 60.7 \\
                &CADA-VAE \cite{Schonfeld2019Generalized} & CVPR'19 & 53.5 & 51.6 & 52.4 & 35.7 & 47.2 & 40.6 & 75.0 & 55.8 & 63.9 \\
                &Li \textit{et al.} \cite{Li2019Rethinking} & ICCV'19 & 47.6 & 47.4 & 47.5 & \textbf{42.8} & 36.3 & 39.3 & 81.4 & 56.4 & 66.7 \\
				&AGZSL \cite{Chou2020Adaptive} & ICLR'21 & 41.4 & 49.7 & 45.2 & 29.9 & 40.2 & 34.3 & 78.9 & \textbf{65.1} & \underline{71.3} \\
                &TDCSS \cite{Feng2022Non} & CVPR'22 & 62.8 & 44.2 & 51.9 & -- & -- & -- & 74.9 & 59.2 & 66.1 \\
                &VGSE \cite{Xu2022VGSE} & CVPR'22 & 24.1 & 45.7 & 31.5 & 25.5 & 35.7 & 29.8 & 45.7 & 66.7 & 54.2 \\
				\midrule
                \multirow{7}{*}{\rotatebox{90}{\textbf{Generation}}}
				&f-CLSWGAN \cite{Xian2018Feature} & CVPR'18 & 43.7 & 57.7 & 49.7 & 36.6 & 42.6 & 39.4 & 57.9 & 61.4 & 59.6 \\
				&LisGAN \cite{Li2019Leveraging} & CVPR'19 & 57.9 & 46.5 & 51.6 & 37.8 & 42.9 & 40.2 & -- & -- & -- \\
				&LsrGAN \cite{Vyas2020Leveraging} & ECCV'20 & 59.1 & 48.1 & 53.0 & 37.7 & 44.8 & 40.9 & 74.6 & 54.6 & 63.0 \\
                &SAGAN \cite{Tang2021ZeroShot} & TNNLS'20 & 59.5 & 45.3 & 51.4 & 29.8 & 44.6 & 35.8  & 84.2 & 55.9 & 67.2 \\
				&CE-GZSL \cite{Han2021Contrastive} & CVPR'21 & 66.8 & 63.9 & 65.3 & \underline{38.6} & \underline{48.8} & \textbf{43.1} & 78.6 & 63.1 & 70.0 \\
				&FREE \cite{Chen2021FREE} & ICCV'21 & 59.9 & 55.7 & 57.7 & 37.2 & 47.4 & \underline{41.7} & 75.4 & 60.4 & 67.1 \\
				\midrule
                \multirow{7}{*}{\rotatebox{90}{\textbf{Attention}}}
				&LFGAA \cite{Liu2019Attribute} & CVPR'19 & \textbf{79.6} & 43.4 & 56.2 & 34.9 & 20.8 & 26.1 & \underline{90.3} & 50.0 & 64.4\\
				&AREN \cite{Xie2019Attentive} & CVPR'19 & \underline{78.7} & 38.9 & 52.1 & 38.8 & 19.0 & 25.5 & \textbf{92.9} & 15.6 & 26.7 \\
				&DAZLE \cite{Huynh2020FineGrained} & CVPR'20 & 56.7 & 59.6 & 58.1 & 24.3 & \textbf{52.3} & 33.2 & 75.7 & 60.3 & 67.1 \\
				&APN \cite{Xu2020Attribute} & NIPS'20 & 69.3 & \underline{65.3} & 67.2 & 34.0 & 41.9 & 37.6 & 78.0 & 56.5 & 65.5 \\
				&GEM-ZSL \cite{Liu2021GoalOriented} & CVPR'21 & 77.1 & 64.8 & \underline{70.4} & 35.7 & 38.1 & 36.9 & 77.5 & 64.8 & 70.6 \\
				\cmidrule(lr){2-12}
                &ALRN & Ours & 77.6 & \textbf{68.2} & \textbf{72.6} & 36.4 & 46.7 & 40.9 & 81.3 & \underline{64.8} & \textbf{72.1} \\
				\bottomrule
		\end{tabular}}
        \caption{GZSL results comparison on CUB, SUN, and AWA2. The best results and second best results are in \textbf{bold} and \underline{underline}, respectively.}
		\label{table:gzsl}
\end{table*}

\begin{table}[ht]
		\centering
		\resizebox{1.0\linewidth}{!}{
			\begin{tabular}{c|c|c|c}
				\toprule
				\textbf{Methods}
				&\textbf{CUB}&\textbf{SUN}&\textbf{AWA2}\\
                \midrule
                SP-AEN \cite{Chen2018Zero} & 55.4 & 59.2 & 58.5 \\
                AREN \cite{Xian2018Feature} & 71.8 & 60.6 & 67.9 \\
				SGMA \cite{Zhu2019SemanticGuided} & 71.0 & - & 68.8 \\
                DAZLE \cite{Huynh2020FineGrained} & 66.0 & 59.4 & 67.9 \\
                APN \cite{Xu2020Attribute} & 72.0 & 61.6 & 58.4 \\
                TransZero \cite{Chen2021TransZero} & \underline{76.8} & \underline{65.6} & \textbf{70.1} \\
                MSDN \cite{Chen2022MSDN} & 76.1 & \textbf{65.8} & \textbf{70.1} \\
                \midrule
                ALRN & \textbf{77.2} & 65.1 & \underline{69.7} \\
				\bottomrule
		\end{tabular}}
        \caption{Per-class Top-1 accuracies comparison under conventional ZSL on CUB, SUN and AWA2 among attention-based methods. The best results and second best results are labeled in \textbf{bold} and \underline{underline}, respectively.}
		\label{table:czsl}
\end{table}

\section{Experiments}\label{experiment}

\subsection{Experimental Setup}
\paragraph{Benchmarks}
We evaluate our model on three popular benchmarks in the ZSL field. Animal with Attributes 2 (AWA2) \cite{Xian2018ZeroShot} is a dataset describing different kinds of animals, which contains 37,322 images belonging to 40 seen classes and 10 unseen classes with each class represented by 50 attributes. Caltech-UCSD-Birds-200-2011 (CUB) \cite{Welinder2010Caltech} consists of 11,788 bird images from 200 classes where each class has 312 attributes. The number of seen and unseen classes for the CUB dataset is set as 150 and 50, respectively. SUN attribute database (SUN) \cite{Patterson2012SUN} is composed of 14,340 images from 717 classes, with 645 seen classes and 72 unseen classes. Each class in SUN can be described by a 102-dimension semantic vector. In terms of data split, we follow the seen/unseen split proposed in \cite{Xian2018ZeroShot} to validate the performance of our method.

\paragraph{Evaluation Metric}
The metrics for conventional ZSL and GZSL are different. For conventional ZSL, the test set only consists of unseen classes, the metric is the per-class Top-1 prediction accuracy for predicting unseen classes, which is denoted as $\bm{T1}$. For GZSL, since the test set contains both seen and unseen classes, two per-class Top-1 prediction accuracy $\bm{S}$ and $\bm{U}$ can be calculated, where $\bm{S}$ represents seen accuracy and $\bm{U}$ means unseen accuracy. To comprehensively evaluate the accuracies on both seen and unseen data, the harmonic mean $\bm{H=(2\times S\times U)/(S + U)}$ is computed to provide a unified evaluation protocol for the GZSL setting. ZSL task seeks balanced predictive performance on seen and unseen classes, so the comparison on harmonic mean $\bm{H}$ is better than simply comparing seen or unseen accuracy.

\begin{figure*}[t]
	\begin{center}
		\includegraphics[width=18cm]{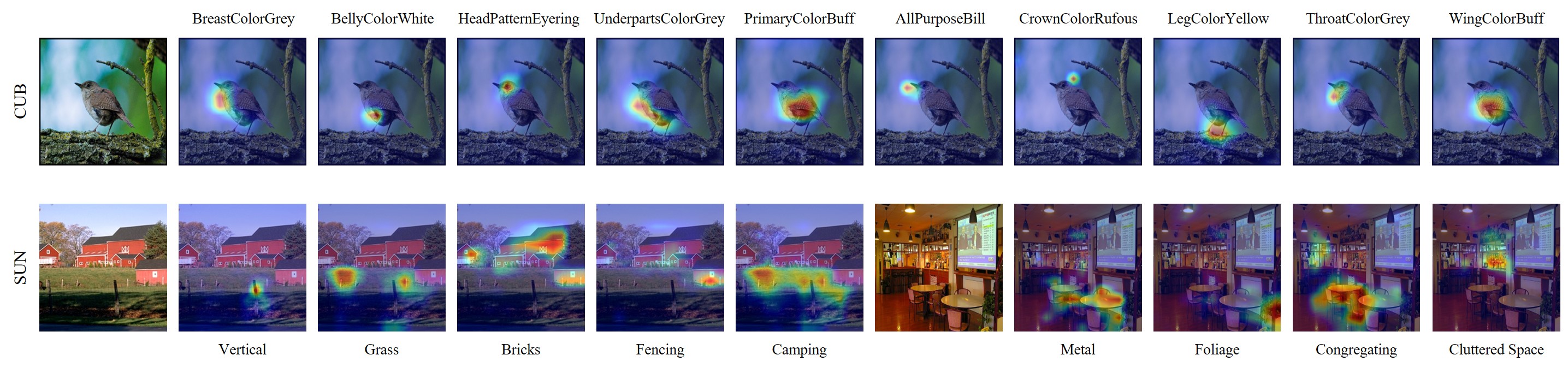}
		\caption{The visualization results of attention maps. For the CUB dataset, one randomly selected image and ten attention maps are shown. For the SUN dataset, two randomly selected images and nine attention maps are exhibited.}
		\label{fig:attention}
	\end{center}
\end{figure*}

\paragraph{Experimental Details}
The backbone of our proposed ALRN is a ResNet101 \cite{He2016Deep} pre-trained on ImageNet-1K. The epoch number $N_{pre}$ for training four different sets of $1\times 1$ convolutional layers, i.e, $f_a$, $f_v$, $f_r$, $f_g$, is set as one for AWA2, five for CUB and SUN. The scaling coefficient $\tau$ is set as 35 for CUB, as well as 20 for SUN and AWA2. The reduction factor $\mu$ is set as 1.5 for SUN, 2.35 for CUB, and 3.9 for AWA2. The weight loss $\lambda$ is set as one for CUB and AWA2, 1.5 for SUN. During the training process, an SGD with momentum optimizer is applied to update parameters, where the learning rate is set as 0.001, the momentum value is set as 0.9, and weight decay is set as 0.00001. Episode-training strategy \cite{Wang2021Region} is adopted which randomly samples images from 16 classes with two images for each class as a mini-batch. We iterate 300 mini-batch in an epoch and train the model 20 epochs for all three datasets.

\subsection{Comparison with Previous Works}

\paragraph{Generalized Zero-Shot Learning}
For the GZSL setting, our ALRN produces the best harmonic mean 72.6\% on CUB, which is 2.2\% higher than the second best result 70.4\% obtained by GEM-ZSL \cite{Liu2021GoalOriented}. Also, our method gets 72.1\% on AWA2, which is 0.8\% higher than the result obtained by AGZSL \cite{Chou2020Adaptive}. For the SUN dataset, the contents of images are complex scenes rather than concrete objects, it is difficult for the model to analyze the detailed information within each image. In such a case, it is challenging for attention-based methods to localize attributes on the SUN dataset. Nevertheless, our ALRN still achieves the highest harmonic mean 40.9\% on the SUN dataset, leading by 3.3\% harmonic mean with the second best harmonic mean produced by APN \cite{Xu2020Attribute}, which is also the most representative method in attention-based methods. In general, benefiting from balanced performance on both seen and unseen classes, our method achieves the highest harmonic mean on two datasets, i.e., CUB and AWA2. Also, the competitive performance of SUN supports the effectiveness of integrating global feature and attribute revision mechanism in our method. 

\paragraph{Conventional Zero-Shot Learning}
As shown in Table \ref{table:czsl}, our ALRN gets the best Top-1 accuracy ($\bm{T1}$) on CUB, where the value of $\bm{T1}$ is 0.4\% higher than the best result obtained by TransZero \cite{Chen2021TransZero}. Also, our method achieves the second best $\bm{T1}$ on AWA2, which is only 0.4\% lower than the highest result produced by TransZero \cite{Chen2021TransZero} and MSDN \cite{Chen2022MSDN}. It is worth mentioning that ALRN gets competitive $\bm{T1}$ on SUN, where it is only 0.7\% lower than the best accuracy produced by MSDN \cite{Chen2022MSDN}, and it surpasses 3.5\% than representative method APN \cite{Xu2020Attribute}. 

\begin{table}[t]
	\centering
	\resizebox{1\linewidth}{!}{
		\begin{tabular}{c|ccc|ccc}
			\toprule
			\multirow{2}{*}{\textbf{Algorithm}} &\multicolumn{3}{c|}{\textbf{SUN}}&\multicolumn{3}{c}{\textbf{AWA2}}\\
			&\rm{$S$} & \rm{$U$} & \rm{$H$} &\rm{$S$} & \rm{$U$} & \rm{$H$}\\
			\midrule
            ALRN \textbf{w/o} ARM \& SCU & 29.2 & \textbf{52.7} & 37.6 & 80.4 & 59.3 & 68.3\\
            ALRN \textbf{w/o} SCU & 30.2 & 51.1 & 38.0 & 79.7 & 60.8 & 69.0\\
			ALRN \textbf{w/o} $\mathcal{L}_{MSE}$ & 32.3 & 49.7 & 39.2 & \textbf{84.9} & 58.3 & 69.1\\
            ALRN \textbf{w/o} global feature & 35.0 & 48.3 & 40.5 & 83.0 & 56.3 & 67.1\\
            ALRN (full) & \textbf{36.4} & 46.7 & \textbf{40.9} & 81.3 & \textbf{64.8} & \textbf{72.1} \\
			\bottomrule
	\end{tabular}}
    \caption{Ablation studies under different model structures and training algorithms in SUN and AWA2. The best results are in \textbf{bold}.}
	\label{table:ablation:structure}
\end{table}

\subsection{Ablation Studies}
\paragraph{Model Structure and Training Algorithm} We conduct ablation studies to demonstrate the effectiveness of each module by training the model under different structures in Table \ref{table:ablation:structure}. It is worth noting that if SCU is deleted, the global and local features will be simply averaged together, and if the ARM is deleted, the output feature of ALRN will directly align with class-level semantics. As we can see from Table \ref{table:ablation:structure}, the removal of the global feature leads to a 0.4\% and 5.0\% harmonic mean degradation on SUN and AWA2, respectively. Analogously, the deletion of SCU leads 2.9\% and 3.1\% harmonic mean reduction on two datasets. Besides the experiments about different model structures, the effectiveness of attribute-level alignment loss, i.e., $\mathcal{L}_{MSE}$ is also worth investigating, where the model only obtains 39.2\% on SUN, and 69.1\% on AWA2. In summary, the ablation study results demonstrates that each network module is indispensable in our approach.

\begin{table}[t]
	\centering
	\resizebox{1\linewidth}{!}{
		\begin{tabular}{c|ccc|ccc}
			\toprule
			\multirow{2}{*}{\textbf{Mechanism}} &\multicolumn{3}{c|}{\textbf{SUN}}&\multicolumn{3}{c}{\textbf{AWA2}}\\
			&\rm{$S$} & \rm{$U$} & \rm{$H$} &\rm{$S$} & \rm{$U$} & \rm{$H$}\\
			\midrule
            ALRN \textbf{w/o} ARM & 29.2 & \textbf{52.7} & 37.6 & 80.4 & 59.3 & 68.3\\
            ALRN \textbf{w/} softmax  & 31.6 & 50.1 & 38.8 & 81.6 & 61.8 & 70.3 \\
            ALRN \textbf{w/} sigmoid & \textbf{36.4} & 46.7 & \textbf{40.9} & 81.3 & \textbf{64.8} & \textbf{72.1} \\
			\bottomrule
	\end{tabular}}
    \caption{Ablation studies under three different revision mechanisms in SUN and AWA2. The best results are in \textbf{bold}.}
	\label{table:ablation:revision}
\end{table}

\paragraph{Revision Mechanism} We also explore the performance under different revision operations in Table \ref{table:ablation:revision}. The model without revision, which is achieved by abandoning ARM, drives the model a 3.3\% and 3.8\% harmonic mean decrease on SUN and AWA2. Then, we replace the sigmoid in Eq. \ref{sigmoid} with softmax, which injects competence between different attributes during the revision weights calculation process. The model with softmax obtains 38.8\% on SUN and 70.3\% on AWA2, which are 2.1\% and 1.8\% lower than the harmonic mean obtained by the model trained with sigmoid. This supports that our proposed attribute-level revision mechanism is more suitable for attention-based approaches.

\subsection{Visulization Results}
\paragraph{Visualization of Attention Maps}
We present the attention visualization results of ALM in Figure \ref{fig:attention}. The ALM performs outstanding ability in locating CUB attributes, where the results can be found in the first row of Figure \ref{fig:attention}. In the second row, we present nine attention maps of two images from the SUN dataset. We can find that ALRN correctly recognizes the wooden stick as vertical, and a large area of grass as camping. Furthermore, our ALRN accurately identifies the place where tables and chairs are gathered as congregating and the bar counter as a cluttered space in the last two attention maps in the second row. In summary, although attributes in the SUN dataset are more semantically related to the image, we can still observe excellent attributes localization skills of our model.

\begin{figure}[t]
\begin{center}
    \includegraphics[width=8.7cm,height=2.8cm]{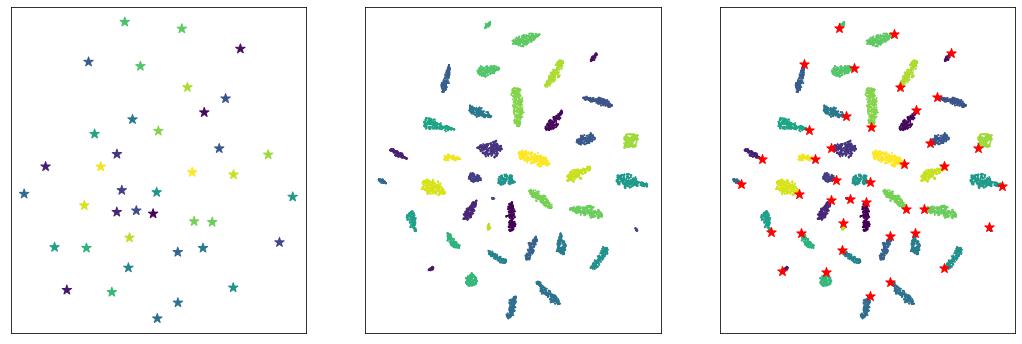} \\
    (a) seen classes \\ \vspace{2mm}
    \includegraphics[width=8.7cm,height=2.8cm]{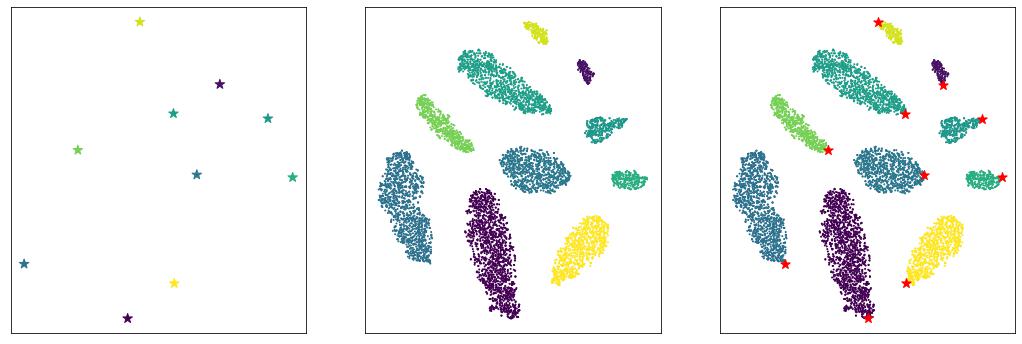} \\
    (b) unseen classes
    \caption{The first two subfigures in each row represent original and revised semantics. The last subfigure combines two previous subfigures, where red stars stand for original semantics.}
    \label{fig:semantics}
\end{center}
\end{figure}

\paragraph{Visualization of Revised Semantics}
For our proposed attribute revision, two concerns may arise. First, the network will produce identical revision weights to all attributes, causing the revised semantics to be the same as the original semantics. Second, attribute revision will break the distribution structure of the original semantics, leading to a less robust model. To deal with these concerns, we visualize the distribution of original and revised semantics to explore the distribution of semantics after conducting attribute revision in Figure \ref{fig:semantics}. Samples from both seen and unseen classes in the test set of AWA2 are fed into the network, then the revised semantics which correspond to the labels are visualized via the t-sne algorithm \cite{Maaten2008tSNE}. As we can see, the revised semantics of different classes exhibit diverging distribution, which starts at the original semantics and diverges in one direction. Also, these revised semantics show good clustering rather than mixed together even for unseen classes. This phenomenon demonstrates our model indeed injects the intra-class variation from the perspective of revising the learning objective and such a revision mechanism is robust enough to prevent semantic features from mixed together and further collapsing.

\begin{figure}[t]
\begin{center}
    \raisebox{1.2cm}{\rotatebox{90}{\footnotesize (1) $N_{pre}$ }}
    \includegraphics[width=4.1cm,height=3.45cm]{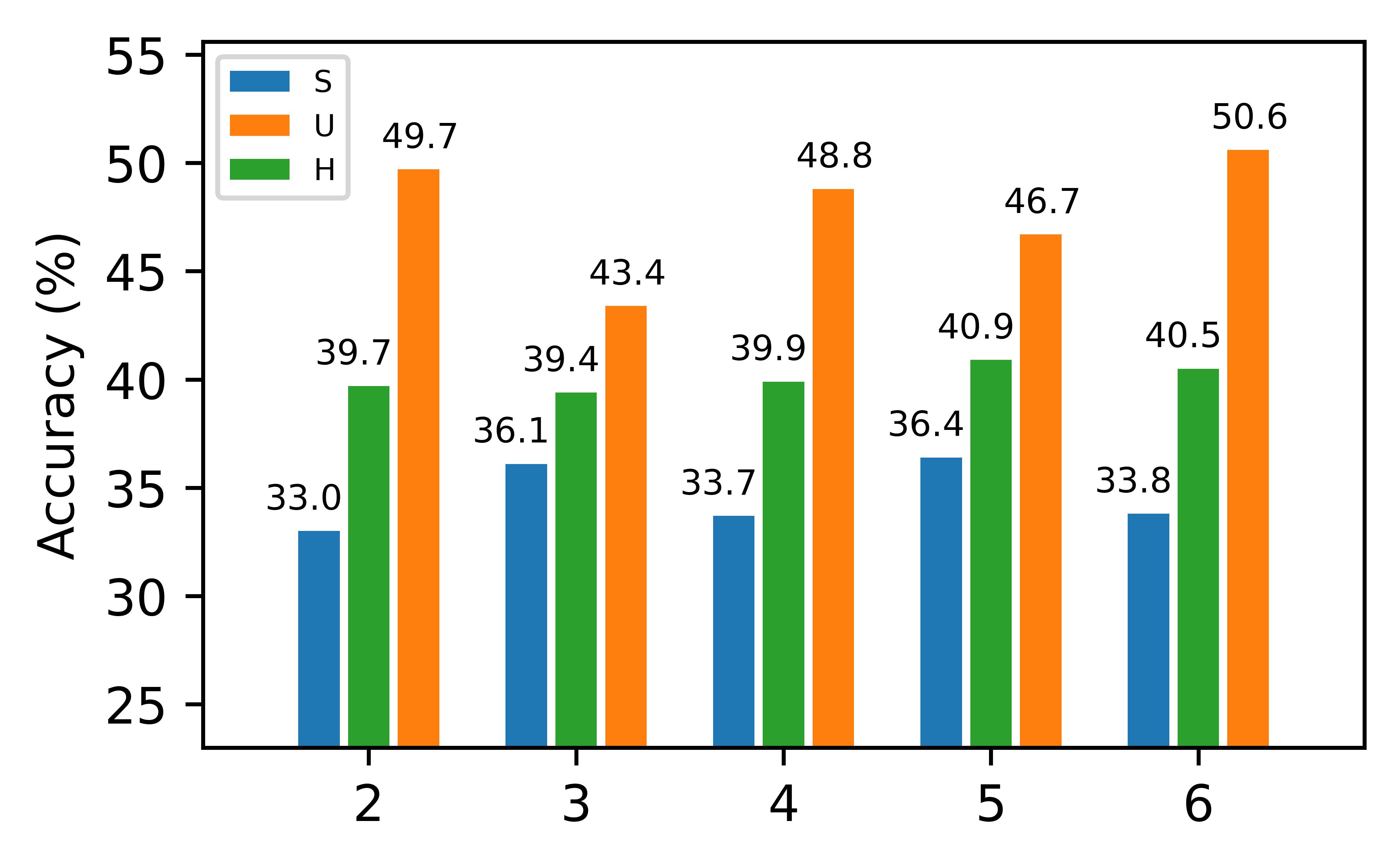}\hspace{-1.5mm}
    \includegraphics[width=4.1cm,height=3.45cm]{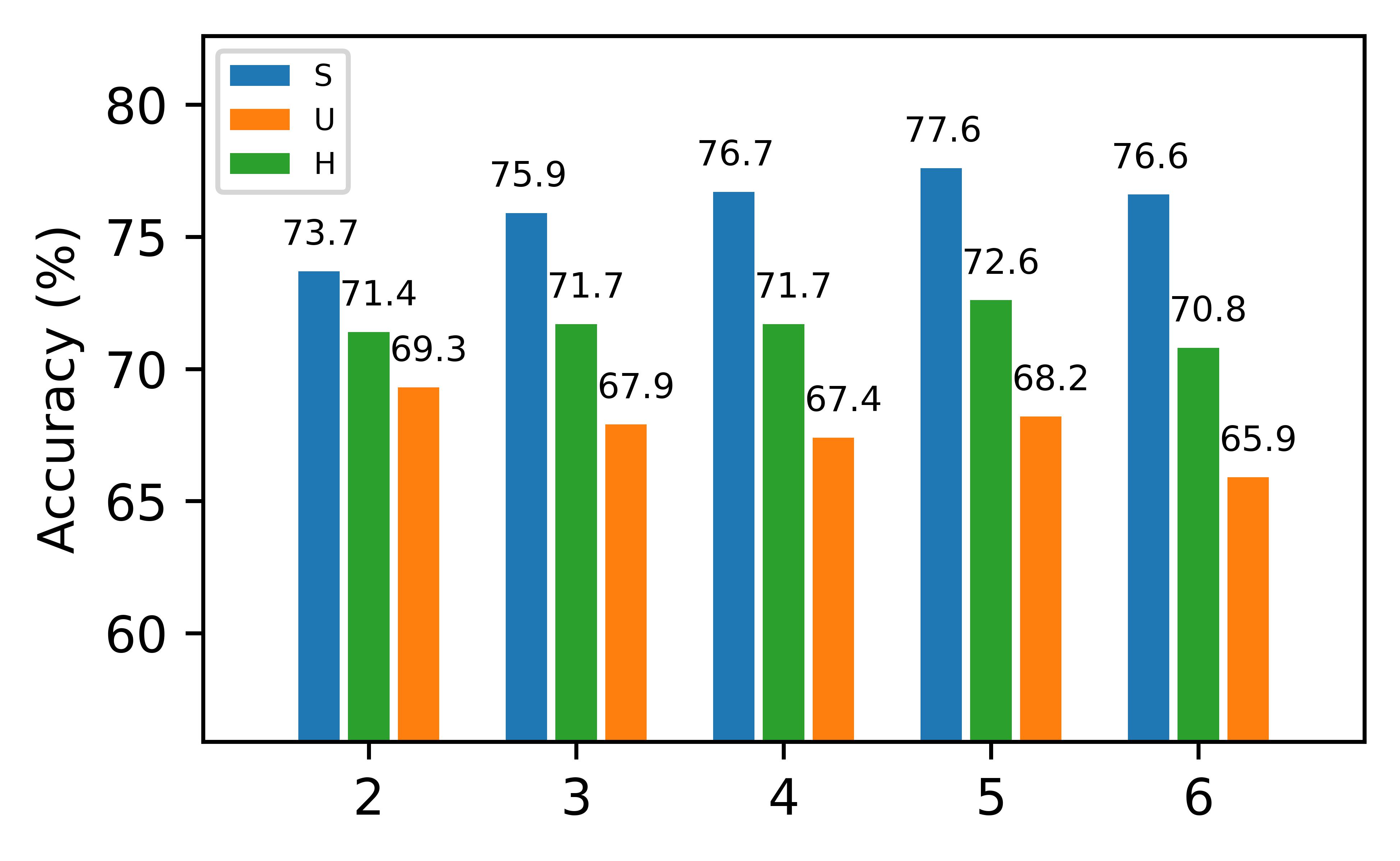} \\
    \raisebox{1.4cm}{\rotatebox{90}{\footnotesize (2) $\tau$ }}
    \includegraphics[width=4.1cm,height=3.45cm]{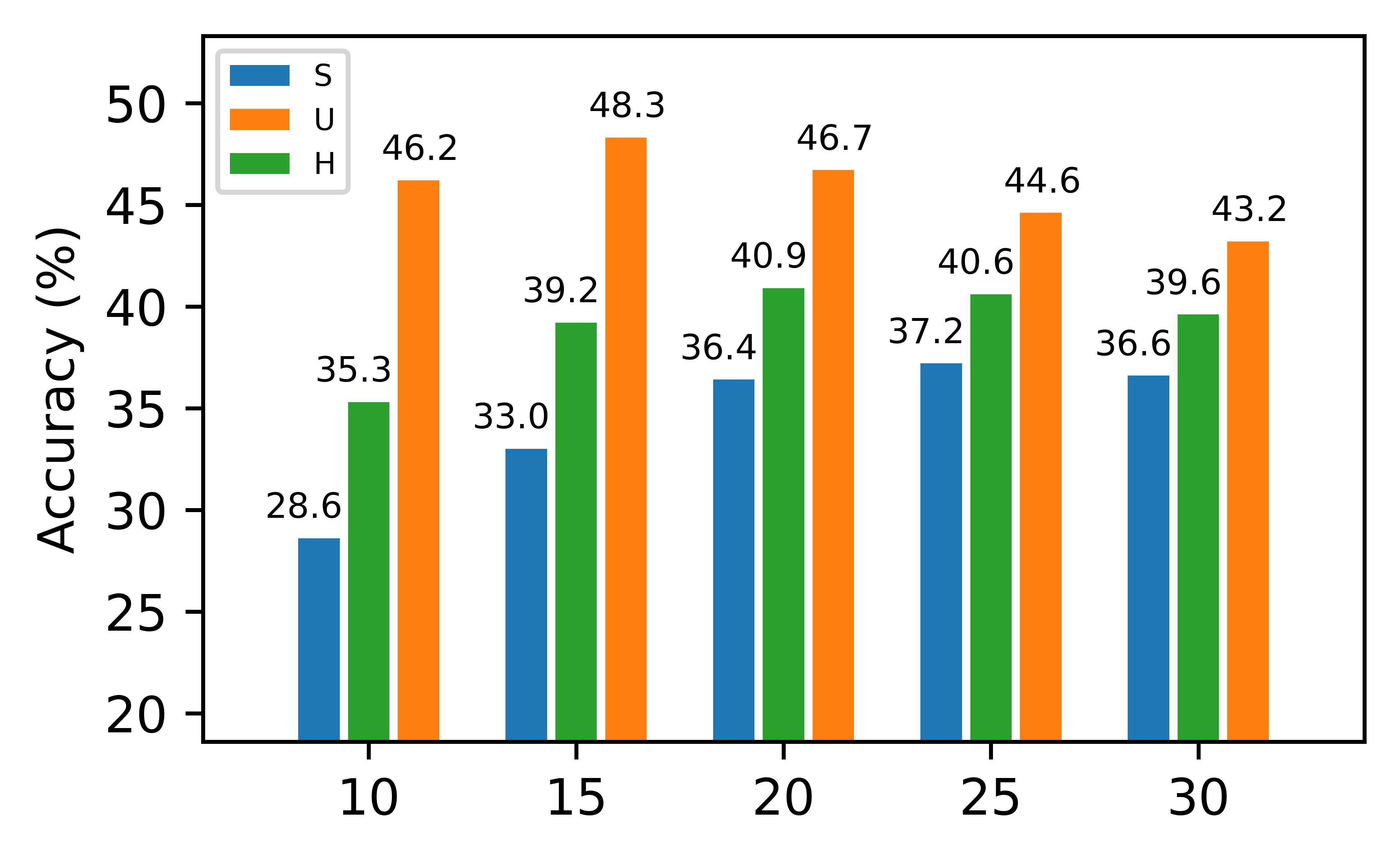}\hspace{-1.5mm}
    \includegraphics[width=4.1cm,height=3.45cm]{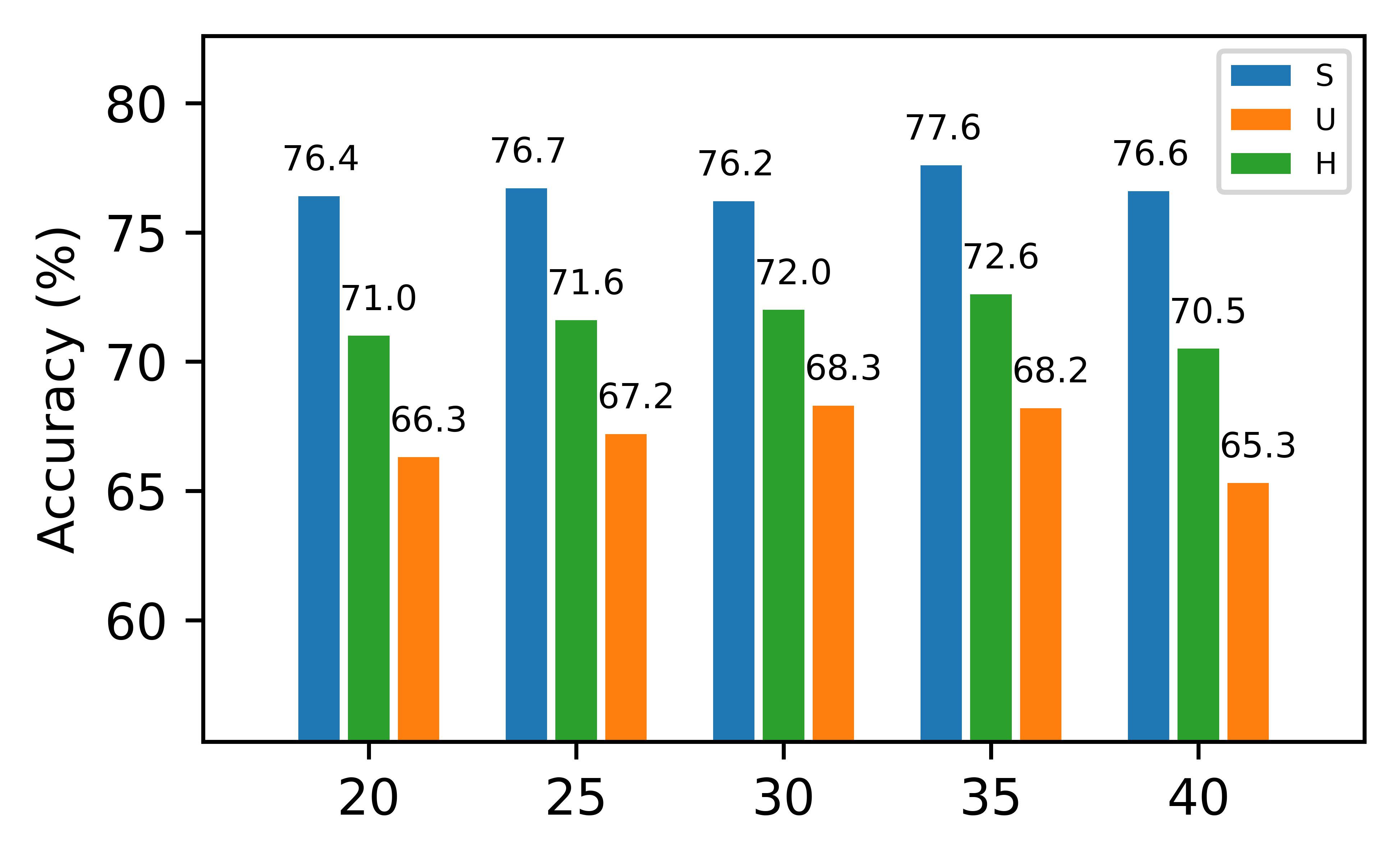} \\
    \raisebox{1.4cm}{\rotatebox{90}{\footnotesize (3) $\lambda$ }}
    \includegraphics[width=4.1cm,height=3.45cm]{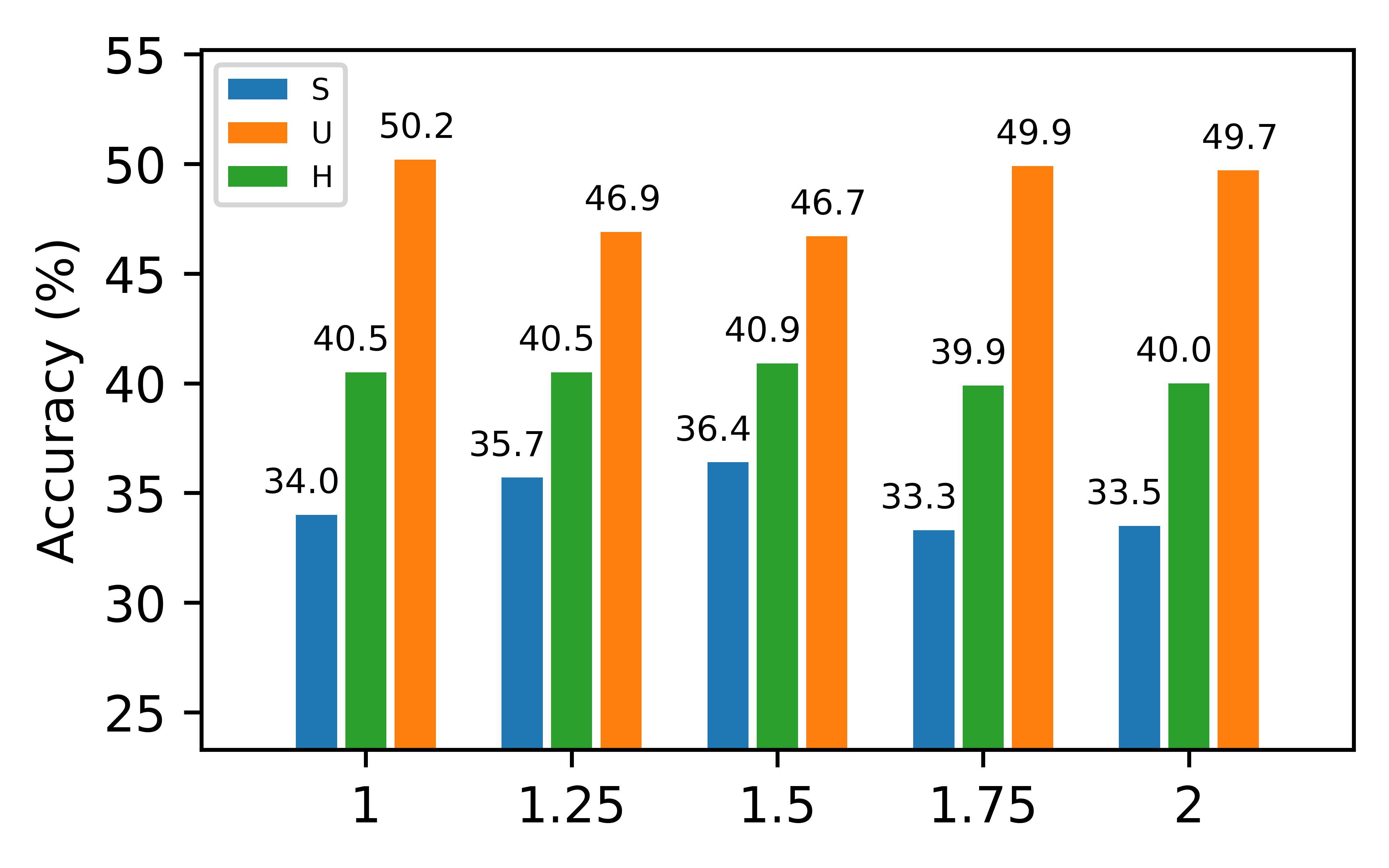}\hspace{-1.5mm}
    \includegraphics[width=4.1cm,height=3.45cm]{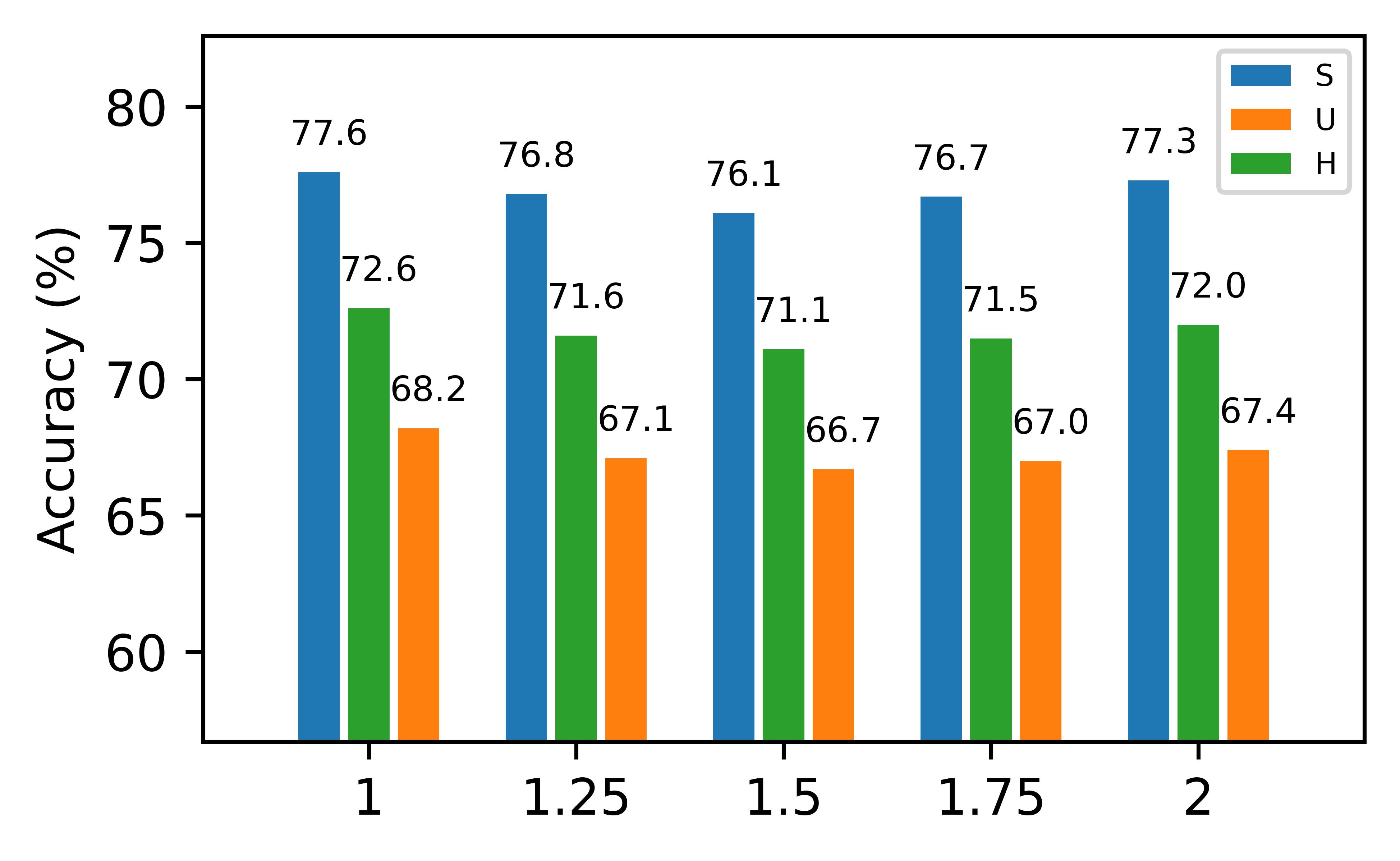} \\
    \hspace{0.5cm} (a) SUN \hspace{3cm} (b) CUB
    \caption{The effect of $N_{pre}$, $\tau$, and $\lambda$ on SUN and CUB.}
    \label{fig:hyperparamter}
\end{center}
\end{figure}

\subsection{Hyper-parameter Analysis}
\paragraph{Effects of Pretraining Epochs} We vary the parameter $N_{pre}$ by selecting one of the exact number in $\{2, 3, 4, 5, 6\}$ on both SUN and AWA2. As we can see from two subfigures in Figure \ref{fig:hyperparamter} (1), the model obtains the highest harmonic mean 72.6\% on CUB and 40.9\% on SUN when $N_{pre}$ is set as 5. This result demonstrates that it is necessary to first adapt four sets of convolutional kernels with the pre-trained backbone, then conduct end-to-end training with the whole network.

\paragraph{Effects of Scaling Coefficient} For the effects of scaling parameter $\tau$, which are exhibited in two subfigures of Figure \ref{fig:hyperparamter} (2), our model gets the highest $H$ value 72.6\% on CUB when $\tau$ is set as 20 and gets the best $H$ value 40.9\% when $\tau$ equals 35. Since the attributes in CUB are local attributes that are easily captured and recognized, a relatively low scaling coefficient is enough to distinguish different classes. However, SUN is a scene dataset where image samples are hard to understand, the model needs a larger scaling coefficient to separate the prediction of different samples.

\paragraph{Effects of Loss Weight} The weight hyper-parameter directly determines how much attribute alignment loss $\mathcal{L}_{MSE}$ is performed in ALRN. From the third row of Figure \ref{fig:hyperparamter}, we find that ALRN performs best on SUN when $\lambda$ is set as 1.5 and gets the highest harmonic mean on CUB when $\lambda$ is set as 1. The difficulty of identifying attributes in SUN requires ALRN to spend more effort on aligning the saliency values of each attribute rather than on the overall semantics. 


\section{Conclusion}\label{conclusion}
In this work, we propose Attribute Localization and Revision Network (ALRN), which tackles the deficiencies of the previous attention-based works that ignore global features and intra-class attribute variation. Experimental results on three widely used benchmarks have shown a significant ability in zero-shot prediction and attribute localization. In future works, we will explore a better way to help the model understand holistic attributes. Also, we will attempt to use vision transformers \cite{Dosovitskiy2020Image} as our backbone to achieve attribute localization and revision.


\bibliographystyle{named}
\bibliography{ijcai23}

\end{document}